\documentclass[11pt]{article}
\usepackage{colacl}
\usepackage{times}
\usepackage{epsf}
\usepackage[latin1]{inputenc}
\usepackage[T1]{fontenc}

\title{A Chart-Parsing Algorithm for Efficient Semantic Analysis}
\author{Pascal Vaillant\\
	ENST/TSI\\
	46, rue Barrault, 75634 Paris cedex 13, France\\
	E-mail: \texttt{vaillant@tsi.enst.fr}}

\begin{document}

\maketitle

\begin{abstract}

In some contexts, well-formed natural language cannot be expected as
input to information or communication systems.  In these contexts, the
use of grammar-independent input (sequences of uninflected semantic
units like e.g. language-independent icons) can be an answer to the
users' needs.  However, this requires that an intelligent system
should be able to interpret this input with reasonable accuracy and in
reasonable time.  Here we propose a method allowing a purely
semantic-based analysis of sequences of semantic units.  It uses an
algorithm inspired by the idea of ``chart parsing'' known in Natural
Language Processing, which stores intermediate parsing results in
order to bring the calculation time down.

\end{abstract}

\section*{Introduction}

As the mass of international communication and exchange increases,
icons as a mean to cross the language barriers have come through in
some specific contexts of use, where language independent symbols
are needed (e.g. on some machine command buttons). The renewed
interest in iconic communication has given rise to important
works in the field of Design \cite{aicher-krampen96,dreyfuss84,ota93},
on reference books on the history and development of the matter
\cite{frutiger91,liungman95,sassoon-gaur97}, as well as newer
studies in the fields of Human-Computer Interaction and Digital
Media \cite{yazdani-barker00} or Semiotics \cite{vaillant99}.

We are here particularly interested in the field of Information
Technology.  Icons are now used in nearly all possible areas of human
computer interaction, even office software or operating systems.
However, there are contexts where richer information has to be
managed, for instance: Alternative \& Augmentative Communication
systems designed for the needs of speech or language impaired people,
to help them communicate (with icon languages like Minspeak, Bliss,
Commun-I-Mage); Second Language Learning systems where learners have
a desire to communicate by themselves, but do not master the structures
of the target language yet; Cross-Language Information Retrieval
systems, with a visual symbolic input.

In these contexts, the use of icons has many advantages: it makes no
assumption about the language competences of the users, allowing
impaired users, or users from a different linguistic background (which
may not include a good command of one of the major languages involved
in research on natural language processing), to access the systems; it
may trigger a communication-motivated, implicit learning process,
which helps the users to gradually improve their level of literacy in
the target language.  However, icons suffer from a lack of expressive
power to convey ideas, namely, the expression of {\em abstract
relations between concepts} still requires the use of linguistic
communication.

An approach to tackle this limitation is to try to ``analyse''
sequences of icons like natural language sentences are parsed, for
example.  However, icons do not give grammatical information as clues
to automatic parsers.  Hence, we have defined a method to interpret
sequences of icons by implementing the use of ``natural'' semantic
knowledge.  This method allows to build knowledge networks from icons
as is usually done from text.

The analysis method that will be presented here is logically
equivalent to the parsing of a dependency grammar with no locality
constraints.  Therefore, the complexity of a fully recursive parsing
method grows more than exponentially with the length of the input.
This makes the reaction time of the system too long to be acceptable
in normal use.  We have now defined a new parsing algorithm which
stores intermediate results in ``charts'', in the way chart parsers
\cite{earley70} do for natural language.

\section{Description of the problem}

Assigning a {\em signification} to a sequence of information items
implies building conceptual relations between them.  Human linguistic
competence consists in manipulating these dependency relations: when
we say that the cat drinks the milk, for example, we perceive that
there are well-defined conceptual connections between `cat', `drink',
and `milk'---that `cat' and `milk' play given roles in a given
process.  Symbolic formalisms in AI \cite{sowa84} reflect this
approach. Linguistic theories have also been developed specifically
to give account of these phenomena \cite{tesniere59,kunze75,melcuk88},
and to describe the transition between semantics and various levels
of syntactic description: from deep syntactic structures which
actually reflect the semantics contents, to the surface structure
whereby messages are put into natural language.

Human natural language reflects these conceptual relations in its
messages through a series of linguistic clues. These clues, depending
on the particular languages, can consist mainly in word ordering in
sentence patterns (``syntactical'' clues, e.g. in English, Chinese, or
Creole), in word inflection or suffixation (``morphological'' clues,
e.g. in Russian, Turkish), or in a given blend of both (e.g. in
German).  {\em Parsers} are systems designed to analyse natural
language input, on the base of such clues, and to yield a
representation of its informational contents.

\begin{center}
\epsfxsize=82mm
\makebox[79mm][c]{\epsffile{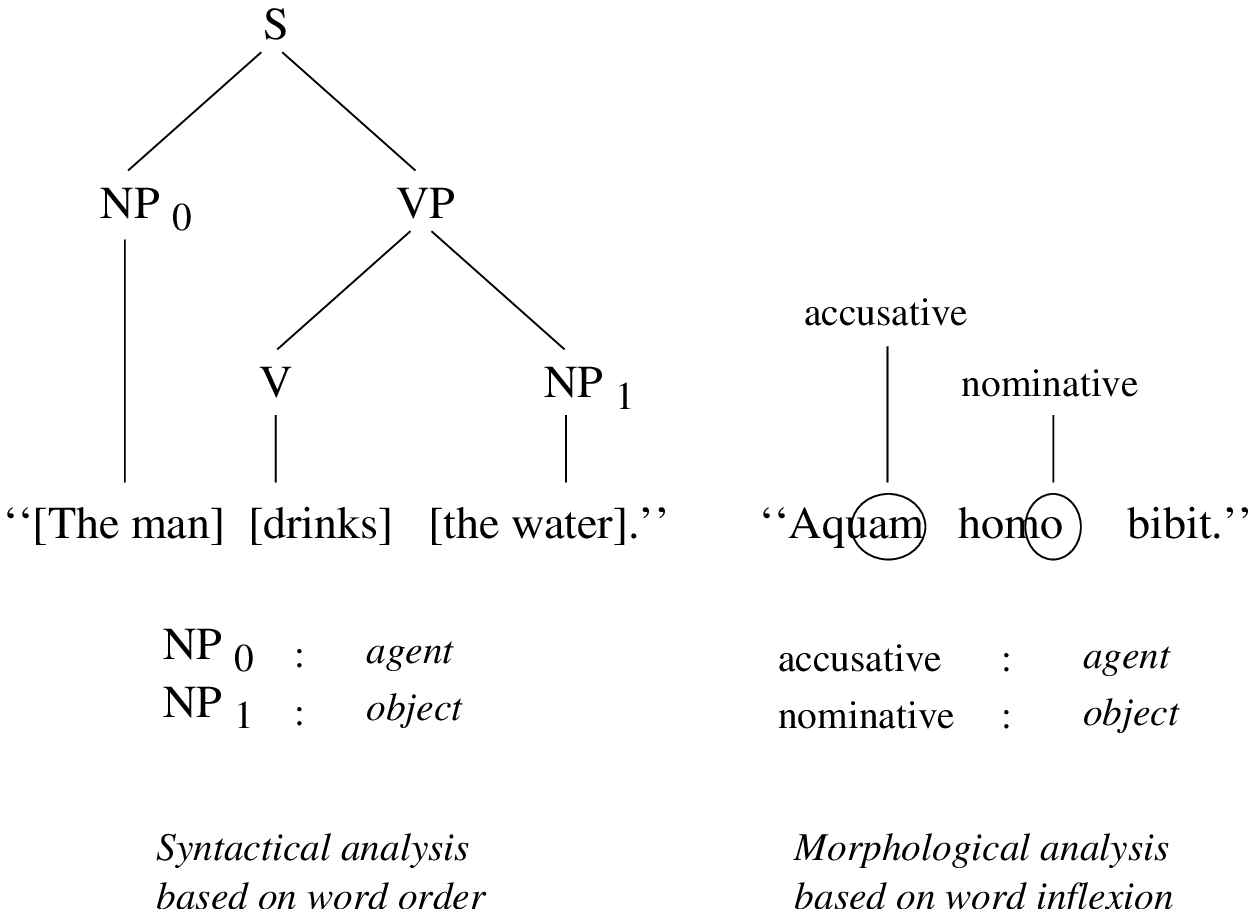}}
\end{center}

In contexts where icons have to be used to convey complex meanings,
the problem is that morphological clues are of course not available,
when at the same time we cannot rely on a precise sentence pattern.

We thus should have to use a parser based on computing the
dependencies, such as some which have been written to cope with
variable-word-order languages \cite{covington90}. However, since no
morphological clue is available either to tell that an icon is, e.g.,
accusative or dative, we have to rely on semantic knowledge to guide
role assignment. In other words, an icon parser has to know that
drinking is something generally done by living beings on liquid
objects.

\section{The semantic analysis method}
\label{method}

The icon parser we propose performs semantic analysis of input
sequences of icons by the use of an algorithm based on
best-unification: when an icon in the input sequence has a
``predicative'' structure (it may become the head of at least one
dependency relation to another node, labeled ``actor''), the other
icons around it are checked for compatibility. Compatibility is
measured as a unification score between two sets of feature
structures: the intrinsic semantic features of the candidate actor,
and the ``extrinsic'' semantic features of the predicative icon
attached to a particular semantic role (i.e. the properties
``expected'' from, say, the agent of {\em kiss}\,, the direct object
of {\em drink}\,, or the concept qualified by the adjective {\em
fierce}\,).

The result yielded by the semantic parser is the graph that maximizes
the sum of the compatibilities of all its dependency relations. It
constitutes, with no particular contextual expectations, and given the
state of world knowledge stored in the iconic database in the form of
semantic features, the ``best'' interpretation of the users' input.

The input is a sequence of icons $s_{1}$, $s_{2}$, \dots{} $s_{n}$,
each of which has a set of intrinsic features:

${\cal IF}(s_{i}) = F_{i}$

\noindent
(where $F_{i}$ is a set of simple Attribute-Value semantic features,
used to represent intrinsic features of the concept\/---\/like
\{\verb"<human,+1>",\verb"<male,+1>"\} for {\em Daddy}).

Some of the symbols also have selectional features, which, if grouped
by case type, form a case structure:

${\cal CS}(s_{i}) =
\{\langle{}c_{1},F_{i1}\rangle{},\langle{}c_{2},F_{i2}\rangle{},
\dots{} \langle{}c_{n},F_{in}\rangle{}\}$

\noindent
(where each of the $n$ $c_{j}$ is a case type such as {\em agent},
{\em object}, {\em goal}..., and each $F_{ij}$ a set of simple
Attribute-Value semantic features, used to determine what features
are {\em expected} from a given case-filler---e.g.
\verb"<human,+1>" is a feature that the {\em agent} of the verb
{\em write} should possess).

Every couple $\langle{}c_{j},F_{ij}\rangle{}$ present in the case
structure means that $F_{ij}$ is a set of Attribute-Value couples
which are attached to $s_{i}$ as selectional features for the case
$c_{j}$:

${\cal SF}(s_{i},c_{j}) = F_{ij}
\Longleftrightarrow{}
\langle{}c_{j},F_{ij}\rangle{} \in {\cal CS}(s_{i})$

For example, we can write:

${\cal SF}$({\em write},{\em agent})$ = $\{\verb"<human,+1>"\}

The {\em semantic compatibility} is the value we seek to maximize to
determine the best assignments.

1. At the feature level (compatibility between two features), it is
defined so as to ``match'' extrinsic and intrinsic features.  This
actually includes a somehow complex definition, taking into account
the modelling of conceptual inheritance between semantic features; but
for the sake of simplicity in this presentation, we may assume that
the semantic compatibility at the semantic feature level is defined as
in Eq.~\ref{c-feature-level}, which would be the case for a ``flat''
ontology\footnote{The difference in computing time may be neglected in
the following reasoning, since the actual formula taking into account
inheritance involves a maximum number of computing steps depending on
the depth of the semantic features ontology, which does not vary
during the processing.}.

\begin{table*}
\begin{equation}
\begin{array}{llrl}
{\cal C}(\langle{}a_{1},v_{1}\rangle{},\langle{}a_{2},v_{2}\rangle{})
& = & 0 & \mbox{if $a_{1} \neq a_{2}$} \\
{\cal C}(\langle{}a,v_{1}\rangle{},\langle{}a,v_{2}\rangle{})
& = & +1 & \mbox{if $v_{1}$ and $v_{2}$ are equal integers} \\
& & -1 & \mbox{if $v_{1}$ and $v_{2}$ are distinct integers} \\
& & v_{1}.v_{2} & \mbox{if one of the values is real}
\end{array}
\label{c-feature-level}
\end{equation}
\end{table*}

2. At the feature structure level, i.e. where the semantic contents of
icons are defined, semantic compatibility is calculated between two
homogeneous sets of Attribute-Value couples: on one side the
selectional features attached to a given case slot of the predicate
icon---stripped here of the case type---, on the other side the
intrinsic features of the candidate icon.

The basic idea here is to define the compatibility as the sum of
matchings in the two sets of attribute-value pairs, in ratio to the
number of features being compared to.  It should be noted that
semantic compatibility is not a symmetric norm: it has to measure how
good the candidate actor fills the expectations of a given predicative
concept in respect to one of its particular cases.  Hence there is a
{\em filtering} set (${\cal SF}$) and a {\em filtered} set (${\cal
IF}$), and it is the cardinal of the filtering set which is used as
denominator:

\vspace{-2ex}

{\setlength\arraycolsep{2pt}
\begin{eqnarray}
{\cal C}({\cal IF},{\cal SF}) & = & {\cal C}(\{f_{11},\dots{},f_{1m}\},\{f_{21},\dots{},f_{2n}\})\nonumber\\
 & = & \frac{\sum_{j \in [1,n]}{\sum_{i \in [1,m]}{{\cal C}(f_{1i},f_{2j})}}}{n}
\label{c-feature-structure-level}
\end{eqnarray}}

\vspace{-1ex}

\noindent
(where the $f_{1i}$ and the $f_{2j}$ are simple features of the form
$\langle{}a_{1i},v_{1i}\rangle{}$ and
$\langle{}a_{2j},v_{2j}\rangle{}$, respectively).

A threshold of acceptability is used to shed out improbable associations
without losing time.

Even with no grammar rules, though, it is necessary to take into
account the distance between two icons in the sequence, which make it
more likely that the actor of a given predicate should be just before
or just after it, than four icons further, out of its context. Hence
we also introduce a ``fading'' function, to weight the virtual
semantic compatibility of a candidate actor to a predicate, by its
actual distance to the predicate in the sequence:

\vspace{-2ex}

{\setlength\arraycolsep{2pt}
\begin{equation}
{\cal V}(s_{i},c_{j},s_{k}) = D(s_{i},s_{k}).{\cal C}({\cal IF}(s_{k}),{\cal SF}(s_{i},c_{j}))
\label{weighted-value}
\end{equation}}

\vspace{-1ex}

\noindent
where:

\({\cal V}(s_{i},c_{j},s_{k})\) is the value of the
assignment of candidate icon $s_{k}$ as filler of the role $c_{j}$
of predicate $s_{i}$;

$D$ is the fading function (decreasing from 1 to 0
when the distance between the two icons goes from 0 to $\infty{}$);

and \({\cal C}({\cal IF}(s_{k}),{\cal SF}(s_{i},c_{j}))\) the
(virtual) semantic compatibility of the intrinsic features of $s_{k}$
to the selectional features of $s_{i}$ for the case $c_{j}$, with no
consideration of distance (as defined in
Eq.~\ref{c-feature-structure-level}).

3. Eventually a global {\bf assignment} of actors (chosen among those
present in the context) to the case slots of the predicate, has to be
determined.  An assignment is an application of the set of icons
(other than the predicate being considered) into the set of cases of
the predicate.

The semantic compatibility of this global assignment is
defined as the sum of the {\em values} (as defined in
Eq.~\ref{weighted-value}) of the individual case-filler allotments.

4. For a sequence of icon containing more than one predicative symbol,
the calculus of the assignments is done for every one of them. A
global {\bf interpretation} of the sequence is a set of assignments
for every predicate in the sequence.

\section{Complexity of a recursive algorithm}
\label{complexity-recursive}

In former works, this principle was implemented by a recursive
algorithm (purely declarative {\sc Prolog}).  Then, for a sequence of
$N$ concepts, and supposing we have the (mean value of) $V$ (valency)
roles to fill for every predicate, let us evaluate the time we need to
compute the possible interpretations of the sequence, when we are in
the worst case, i.e. the $N$ icons are all predicates.

1. For every assignment, the number of semantic compatibility
values corresponding to a single role/filler allotment, on an
\( \langle{}actor,candidate\rangle{} \) couple (i.e. at the feature
structure level, as defined in Eq.~\ref{c-feature-structure-level})
is: \( (N-1) \times{} V \).

2. For every icon, the number of possible assignments is:

\vspace{-2ex}

\begin{equation}
^{N-1}\/P\/_{V} = \frac{(N-1)!}{(N-1-V)!}
\end{equation}

\vspace{-1ex}

(we suppose that \(N-1 > V\), because we are only interested in what
happens when $N$ becomes big, and $V$ typically lies around 3).

3. For every assignment, the \(N-1\) allotment possibilities for
the first case are computed only once.  Then, for every possibility of
allotment of the first case, the \(N-1\) possibilities for the second
case are recomputed---hence, there are \((N-1)^{2}\) calculations of
role/filler allotment scores for the second case.  Similarly, every
possible allotment for the third case is recomputed for every possible
choice set on the first two cases---so, there are \((N-1)^{3}\)
computations on the whole for the third case. This goes on until the
$V^{th}$ case.

In the end, for one single assignment, the number of times a
case/filler score has been computed is \( \sum_{k=1}^{V}{(N-1)^{k}} \).

\noindent
Then, to compute all the possible interpretations:

1. Number of times the system computes every possible assignment
of the first icon: 1.

2. Number of times the system computes every possible assignment
of the second icon: \( ^{N-1}\/P\/_{V} \) (once for every assignment of
the first icon, backtracking every time---still supposing we are in
the worst case, i.e. all the assignments pass over the acceptability
threshold).

3. Number of times the system computes every possible assignment
of the third icon: \( ^{N-1}\/P\/_{V} \times{} ^{N-1}\/P\/_{V} \) (once for
every possible assignment of the second icon, each of them being
recomputed once again for every possible assignment of the first
icon). ( \dots{} )

4. Number of times the system computes every possible assignment of
the $N^{th}$ icon: \( ( ^{N-1}\/P\/_{V} )^{N-1} \).

5. Number of assignments computed on the whole: every assignment
of the first icon (there are \( ^{N-1}\/P\/_{V} \) of them) is computed
just once, since it is at the beginning of the backtracking chain;
every assignment of the second icon is computed \( ^{N-1}\/P\/_{V} \)
times for every assignment of the first icon, so \( (^{N-1}\/P\/_{V})^{2} \)
times, \dots{} every assignment of the $N^{th}$ icon is computed
\( (^{N-1}\/P\/_{V})^{N} \) times.

Total number of assignment calculations: \hfill\linebreak[4]
\( \sum_{k=1}^{N}{(^{N-1}\/P\/_{V})^{k}} \).

\vspace{1ex}

6. Every calculation of an assignment value involves, as we have
seen, \( \sum_{k=1}^{V}{(N-1)^{k}} \) calculations of a semantic
compatibility at a feature structure level.  So, totally, for the
calculation of all possible interpretations of the sentence, the
number of such calculations has been:

\vspace{-1ex}

\[ \sum_{k=1}^{V}{(N-1)^{k}} \times{} \sum_{k=1}^{N}{(^{N-1}\/P\/_{V})^{k}} \]

\vspace{-1ex}

7. Lastly, the final scoring of every interpretation involves summing
the scores of the $N$ assignments, which takes up $N-1$ elementary
(binary) sums.  This sum is computed every time an interpretation is
set, i.e. every time the system reaches a leaf of the choice tree,
i.e. every time an assignment for the $N^{th}$ icon is reached, that
is \( (^{N-1}\/P\/_{V})^{N} \) times.  So, there is an additional
computing time which also is a function of $N$, namely, expressed
in number of elementary sums:

\vspace{-2ex}

\[ (N-1) \times{} (^{N-1}\/P\/_{V})^{N} \]

\vspace{-1ex}

Hence, if we label $a$ the ratio of the computing time used to
compute the score of a role/filler allotment to the computing time of
an elementary binary sum\footnote{$a$ is a constant in relation to
$N$: the computation of the semantic compatibility at the feature
structure level, defined in Eq.~\ref{c-feature-structure-level},
roughly involves $n\times{}m$ computations of the semantic
compatibility at the feature level, defined in
Eq.~\ref{c-feature-level} ($n$ being the average number of selectional
features for a given role on a given predicate, and $m$ the average
number of intrinsic features of the entries in the semantic lexicon),
which itself involves a sequence of elementary operations
(comparisons, floating point number multiplication).  It does not
depend on $N$, the number of icons in the sequence.}, the number of
elementary operations involved in computing the scores of the
interpretations of the whole sequence is:

\vspace{-1ex}

\noindent
\begin{equation}
\kern-1ex (N-1).(^{N-1}\/P\/_{V})^{N} + a \sum_{k=1}^{V}{(N-1)^{k}} . \sum_{k=1}^{N}{(^{N-1}\/P\/_{V})^{k}}
\label{eq-complexity-recursive}
\end{equation}

\vspace{-1ex}

\section{The chart algorithm}
\label{chart-algorithm}

To avoid this major impediment, we define a new algorithm which stores
the results of the low-level operations uselessly recomputed at every
backtrack:

\begin{itemize}

\item[a.] The low-level role/filler compatibility values, in a chart
called `{\tt compatibility\_table}'. The values stored here correspond
to the values defined at Eq.~\ref{c-feature-structure-level}.

\item[b.] The value of every assignment, in `{\tt assignments\_table}'.
The values stored here correspond to assignments of multiple case
slots of a predicate, as defined at point 3 of Section~\ref{method};
they are the sum of the values stored at level (a), multiplied by a
fading function of the distance between the icons involved.

\item[c.] The value of the interpretations of the sentence, in `{\tt
interpretations\_table}'. The values stored here correspond to global
interpretations of the sentence, as defined at point 4 of
Section~\ref{method}.

\end{itemize}

With this system, at level (b) (calculation of the values of
assignments), the value of the role/filler couples are re-used from
the compatibility table, and are not recomputed many times. In the
same way, at level (c), the computation of the interpretations' values
by adding the assignments' values does not recompute the assignments
values at every step, but simply uses the values stored in the
assignments table.

Furthermore, the system has been improved for the cases where only
partial modifications are done to the graph, e.g. when the users want
to perform an incremental generation, by generating the graph again at
every new icon added to the end of the sequence; or when they want to
delete one of the icons of the sequence only, optionally to replace it
by another one. In these cases, a great part of the information
remains unchanged. To take this property into account, the system
stores the current sequence and the charts resulting from the parse in
memory, allowing them to be only partially replaced afterwards.

Finally, we have implemented three basic interface functions to be
performed by the parser. The first one implements a full parse, the
second partially re-parses a sequence where new icons have been added,
the third partially re-parses a sequence where icons have been removed.
The three functions can be described as follows.

\paragraph{Parsing from scratch:}

\begin{enumerate}

\item Spot the icons in the new sequence which are potential
predicates (which have a valency frame).

\item Run through the sequence and identify every possible
pair $\langle{}\langle{}$predicate,role$\rangle{}$,candidate$\rangle{}$.

For each one of them, calculate the semantic compatibility

\vspace{-2ex}

\begin{center}

\({\cal C}({\cal IF}(\)candidate\(),{\cal SF}(\)predicate,role\())\).

\end{center}

\vspace{-1ex}

Store all the values found in {\tt compatibility\_table}:

\begin{center}

\begin{tabular}{|lll|l|}\hline
predicate 1 & role 1 & candidate 1 & {\em value}\\
predicate 1 & role 1 & candidate 2 & {\em value}\\
\multicolumn{3}{|c|}{\dots{}} & \\
predicate $k$ & role $V$ & candidate $N$ & {\em value}\\ \hline
\end{tabular}

\end{center}

(and eliminate values under the threshold as soon as they appear).

\item Go through the sequence and identify the set of possible assignments
for each predicate.

For every assignment, compute its score using the values stored in
{\tt compatibility\_table}, and multiplying by the fading coefficients
$D(1)$, $D(2)$, \dots{}

Store the values found in:\\
{\tt assignments\_table} (Tab.~\ref{assignments}).

\begin{table*}
\begin{tabular}{|crclc|l|}\hline
$\langle{}$ & predicate 1 & , & \{ $\langle{}$ role $1$ , candidate $f_{11}(1)$ $\rangle{}$ , \dots{} $\langle{}$ role $V$ , candidate $f_{11}(V)$ $\rangle{}$ \} & $\rangle{}$ & {\em value}\\
$\langle{}$ & predicate 1 & , & \{ $\langle{}$ role $1$ , candidate $f_{12}(1)$ $\rangle{}$ , \dots{} $\langle{}$ role $V$ , candidate $f_{12}(V)$ $\rangle{}$ \} & $\rangle{}$ & {\em value}\\
\multicolumn{5}{|c|}{\dots{}} & \\
$\langle{}$ & predicate $k$ & , & \{ $\langle{}$ role $1$ , candidate $f_{kn}(1)$ $\rangle{}$ , \dots{} $\langle{}$ role $V$ , candidate $f_{kn}(V)$ $\rangle{}$ \} & $\rangle{}$ & {\em value}\\ \hline
\end{tabular}
\caption[assignments]{Assignments Table}
\label{assignments}
\end{table*}

\item Calculate the list of all the possible interpretation (1
interpretation is 1 sequence of assignments). Store them along with
their values in {\tt interpretations\_table}.

\end{enumerate}

\paragraph{Add a list of icons to the currently stored sequence:}

\begin{enumerate}

\item Add the icons of {\em list of icons} to the currently stored sequence.

\item For every pair $\langle{}\langle{}$predicate,role$\rangle{}$,candidate$\rangle{}$.

where either the predicate, or the candidate, is a new icon (is a member
of {\em list of icons}), calculate the value of

\vspace{-2ex}

\begin{center}

\({\cal C}({\cal IF}(\)candidate\(),{\cal SF}(\)predicate,role\())\).

\end{center}

\vspace{-1ex}

and store the value in:\\
{\tt compatibility\_table}.

\item Calculate the new assignments made possible by the new icons from
{\em list of icons}:

\begin{itemize}

\item the assignments of new predicates;

\item for every predicate already present in the sequence before, the assignments
      where at least one of the roles is allotted to one of the icons of
      {\em list of icons}.

\end{itemize}

For each of them, calculate its value, and store it in {\tt assignments\_table}.

\item Recompute the table of interpretations totally (no get-around).

\end{enumerate}

\paragraph{Remove a list of icons from the currently stored sequence:}

\begin{enumerate}

\item Remove the icons of {\em list of icons} from the sequence stored in
memory.

\item Remove the entries of {\tt compatibility\_table} or {\tt
assignments\_table} involving at least one of the icons of {\em list of icons}.

\item Recompute the table of interpretations.

\end{enumerate}

\section{Complexity of the chart algorithm}

First, let us evaluate the complexity of the algorithm presented in
Section~\ref{chart-algorithm} assuming that only the first interface
function is used (parsing from scratch every time a new icon is
added to the sequence).

In the worst case: the $N$ icons are all predicates; no possible
role/filler allotment in the whole sequence is below the threshold of
acceptability.

\begin{itemize}

\item For every predicate, every combination between one single role
and one single other icon in the sequence is evaluated: there are
\( (N-1) \times{} V \) such possible couples
\( \langle{}actor,candidate\rangle{} \).

\item Since there are (worst case) $N$ predicates, there are
\( N \times{} (N-1) \times{} V \) such combinations to compute for
the whole sequence, in order to fill the compatibility table.

\item After the compatibility table has been filled, its values are
used to compute the score of every possible assignment (of surrounding
icons) for every predicate (to its case roles).  Computing the score
of an assignment involves summing $V$ values of the compatibility
table, multiplied by a value of the fading function $D$, typically for
a small integer.  Thus, for every line in the assignments table
(Table~\ref{assignments}), the computing time is constant in respect
to $N$.

\item For every predicate, there are

\vspace{-2ex}

\[ ^{N-1}\/P\/_{V} = \frac{(N-1)!}{(N-1-V)!} \]

\vspace{-1ex}

possible assignments (see Section~\ref{complexity-recursive}). Since
there are $N$ predicates, there is a total number (in the worst case)
of \( N \times{} ^{N-1}\/P\/_{V} \) different possible assignments,
i.e. different lines to fill in the assignments table.  So, the time
to fill the assignment table in relation to $N$ is \(N!/(N-1-V)!\)
multiplied by a constant factor.

\item After the assignments table has been filled, its values are
used to compute the score of the possible interpretations of the
sentence.  The computation of the score of every single interpretation
is simply a sum of scores of assignments: since there possibly are
$N$ predicates, there might be up to $N$ figures to sum to compute
the score of an interpretation.

\item An interpretation is an element of the cartesian product
of the sets of all possible assignments for every predicate. Since
every one of these sets has \( ^{N-1}\/P\/_{V} \) elements, there
is a total number of

\vspace{-2ex}

\[ (^{N-1}\/P\/_{V})^{N} = \frac{(N-1)!^{N}}{(N-1-V)!^{N}} \]

\vspace{-1ex}

interpretations to compute.  As each computation might involve $N-1$
elementary sums (there are $N$ figures to sum up), we may conclude that
the time to fill the interpretations table is in a relation to $N$
which may be written so: $(N-1) \times{} (^{N-1}\/P\/_{V})^{N}$.

\item In the end, the calculation time is not the product, but the
sum, of the times used to fill each of the tables.  So, if we label
$a$ and $b$ two constants, representing, respectively, the ratio of
the computing time used to get the score of an elementary role/filler
allotment to the computing time of an elementary binary addition, and
the ratio of the computing time used to get the score of an assignment
from the scores of the role/filler allotments (adding up $V$ of them,
multiplied by values of the $D$ function), to the computing time of an
elementary binary addition, the total computing time for calculating
the scores of all possible interpretations of the sentence is:

\end{itemize}

\vspace{-2ex}

\noindent
\begin{equation}
(N-1).(^{N-1}\/P\/_{V})^{N} + aVN(N-1) + bN(^{N-1}\/P\/_{V})
\label{eq-complexity-chart}
\end{equation}

\vspace{-1ex}

\section{Discussion}

We have presented a new algorithm for a completely semantic parse
of a sequence of symbols in a graph-based formalism.  The new algorithm
has a temporal complexity like in Eq.~\ref{eq-complexity-chart}, to be
compared to the complexity of a purely recursive algorithm, like in
Eq.~\ref{eq-complexity-recursive}.

In the worst case, the second function is still dominated by a
function which grows hyperexponentially in relation to
$N$: the number of possible interpretations multiplied by the
time used to sum up the score of an
interpretation\footnote{Namely, \( (N-1).(^{N-1}\/P\/_{V})^{N} \).}.
In practice, the values of the parameters $a$ and $b$ are fairly large,
so this member is still small during the first steps, but it
grows very quickly.

As for the other member of the function, it is hyperexponential in the
case of Eq.~\ref{eq-complexity-recursive}, whereas it is of order \(
bN(^{N-1}\/P\/_{V}) \), i.e. it is $O(N^{V+1})$, in the case of
Eq.~\ref{eq-complexity-chart}.

Practically, to make the semantic parsing algorithm acceptable, the
problem of the hyperexponential growth of the number of
interpretations has to be eliminated at some point.  In the system we
have implemented, a threshold mechanism allows to reject, for every
predicate, the unlikely assignments.  This practically leaves up only
a small maximum number of assignments in the assignments table, for
every predicate---typically 3.  This means that the number of
interpretations is no longer of the order of \( ^{N-1}\/P\/_{V})^{N} \),
but ``only'' of $3^{N}$: it becomes ``simply'' exponential. This
implementation mechanism makes the practical computing time
acceptable when running on an average computer for input sequences of
no more than approximately 15 symbols.

In order to give a comprehensive solution to the problem, future
developments will try to develop heuristics to find out the best
solutions without having to compute the whole list of all possible
interpretations and sort it by decreasing value of semantic
compatibility.  For example, by trying to explore the search space (of
all possible interpretations) from maximum values of the assignments,
it may be possible to generate only the 10 or 20 best interpretations
without having to score all of them to start with.

\end{document}